\documentclass[cameraready]{Interspeech}
\usepackage{multirow}
\usepackage{acronym}
\usepackage{multicol}
\usepackage{adjustbox}
\usepackage{arydshln} 
\usepackage{multirow}
\usepackage{xcolor}
\definecolor{stressA}{RGB}{0,102,204}   
\definecolor{stressB}{RGB}{200,30,30}   

\newcommand{\stressA}[1]{\textcolor{stressA}{\textbf{#1}}}
\newcommand{\stressB}[1]{\textcolor{stressB}{\textbf{#1}}}

\newacro{TTS}{text-to-speech}

\newacro{CAST}{\textbf{C}ontext-\textbf{A}ware \textbf{S}tress \textbf{T}TS}
\newcommand{\bench}{\ac{CAST}}

\title{Knowing What to Stress: A Discourse-Conditioned Text-to-Speech Benchmark}

\author[affiliation={1}]{Arnon}{Turetzky}
\author[affiliation={2}]{Avihu}{Dekel}
\author[affiliation={2}]{Hagai}{Aronowitz}
\author[affiliation={2}]{Ron}{Hoory}
\author[affiliation={1}]{Yossi}{Adi}

\address{
    $^1$ The Hebrew University of Jerusalem, Israel \\
    $^2$ IBM Research, Israel
}

\email{arnon.turetzky@mail.huji.ac.il}


\usepackage{comment}

\begin{document}

\maketitle
\begin{abstract}
Spoken meaning often depends not only on what is said, but also on which word is emphasized. The same sentence can convey correction, contrast, or clarification depending on where emphasis falls. Although modern text-to-speech (TTS) systems generate expressive speech, it remains unclear whether they infer contextually appropriate stress from discourse alone. To address this gap, we present \bench, a benchmark for evaluating context-conditioned word-level stress in TTS. Items are defined as contrastive context pairs: identical sentences paired with distinct contexts requiring different stressed words. We evaluate state-of-the-art systems and find a consistent gap: text-only language models reliably recover the intended stress from context, yet TTS systems frequently fail to realize it in speech. We release the benchmark, evaluation framework, construction pipeline and a synthetic corpus to support future work on context-aware speech synthesis.

\end{abstract}
\section{Introduction}
\begin{figure}[t!]
    \centering
    \includegraphics[width=1\linewidth]{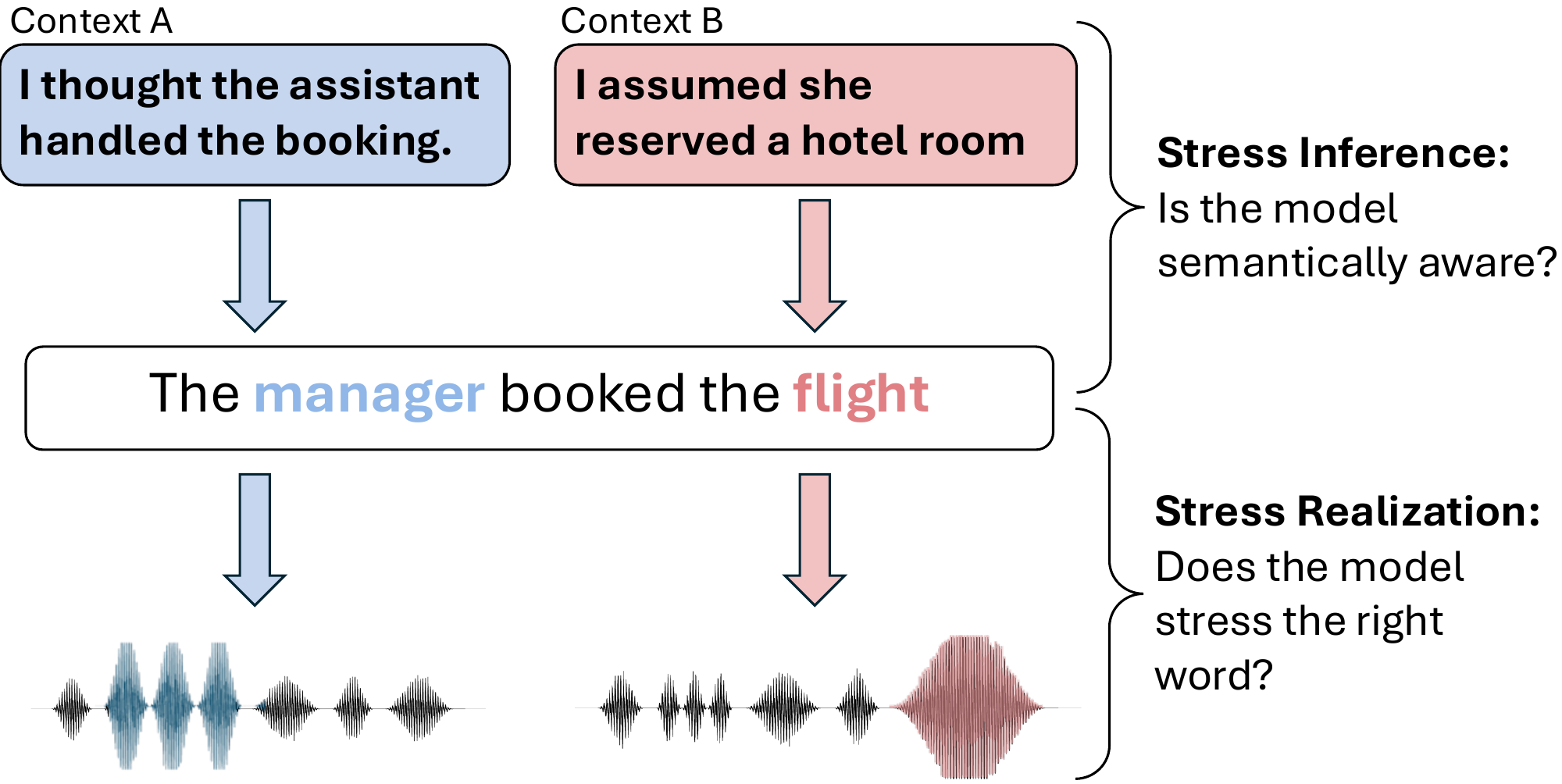}
    \caption{Discourse context determines which word should be stressed in a given sentence.}
    \label{fig:fig1}
\end{figure}

Spoken communication conveys more than lexical content alone. Prosody signals emphasis, contrast, correction, and information structure, shaping how an utterance is interpreted \cite{rooth92,buring2007}. One key aspect of prosody is sentence stress, which refers to emphasis placed on particular words or phrases and can dramatically change meaning for the same written form \cite{Bolinger1972AccentIP, yosha2025stresstest}.
In many conversational settings, the appropriate prosody depends on discourse context rather than the sentence itself. For example, the sentence \textit{``The manager booked the flight''} may emphasize \emph{manager} to correct a mistaken assumption about who performed the action, or emphasize \emph{flight} to clarify what was booked. Although the words remain identical, the intended meaning changes with stress placement. In such cases, lexical stress functions as a marker of contrastive focus, 
as illustrated in Figure~\ref{fig:fig1}.

Recent advances in TTS have enabled high-quality and increasingly expressive speech generation \cite{CosyVoice,chatterboxtts2025,qwen3tts}. Yet it remains unclear whether this natural-sounding prosody reliably reflects discourse-dependent meaning when no explicit emphasis signal is provided. Understanding this requires a controlled evaluation framework that isolates context-dependent stress as a measurable phenomenon.

Prior work has evaluated prosodic phenomena from several perspectives, including stress detection and reasoning \cite{yosha2025whistress,yosha2025stresstest,morrison2024crowdsourced}, emphasis realization with explicit targets \cite{de2024emphassess,de2023prosaudit}, and preservation of stress in spoken translation \cite{de2024emphassess}. While these efforts advance prosody evaluation and modeling, they do not test whether a TTS system infers and shifts stress appropriately across contrasting discourse contexts for identical sentences. We address this gap with the following contributions:
\begin{itemize}
    \item \textbf{\bench}: a benchmark for context-conditioned word-level stress in TTS, where identical sentences are paired with distinct contexts requiring different stressed words, with stress targets defined semantically.
    \item A scalable data generation pipeline used to construct \bench, released to support benchmark regeneration and scaling, with an extended version supporting audio synthesis for broader research on context-aware stress.
    \item A systematic evaluation of state-of-the-art TTS systems, showing that all evaluated systems fail to reliably produce context-appropriate stress in speech across all conditioning modes.
\end{itemize}
In addition, we release a synthetic corpus of $\sim10k$ context-sentence-audio triples generated by the extended pipeline. Unlike the text-only benchmark, 
this extension adds audio synthesis and automatic acoustic stress validation. The corpus is made available to support future research.

\section{Related Work}
\label{sec:related}

Neural TTS systems increasingly support expressive and controllable prosody. Prior work has explored global style embeddings and style tokens to modulate speaking style \cite{wang2018style,skerry2021prosody,lee2021styler}, as well as prosody transfer from reference speech \cite{skerry2018towards}. More recent systems demonstrate accurate realization of word-level stress when it is explicitly specified through control tokens or annotations \cite{CosyVoice,korotkova2024word}. While these approaches improve prosodic controllability, they assume that the appropriate stress target is provided at inference time and do not address the problem of inferring stress from discourse context.
\begin{figure*}[t!]
    \centering
    \includegraphics[width=1\linewidth]{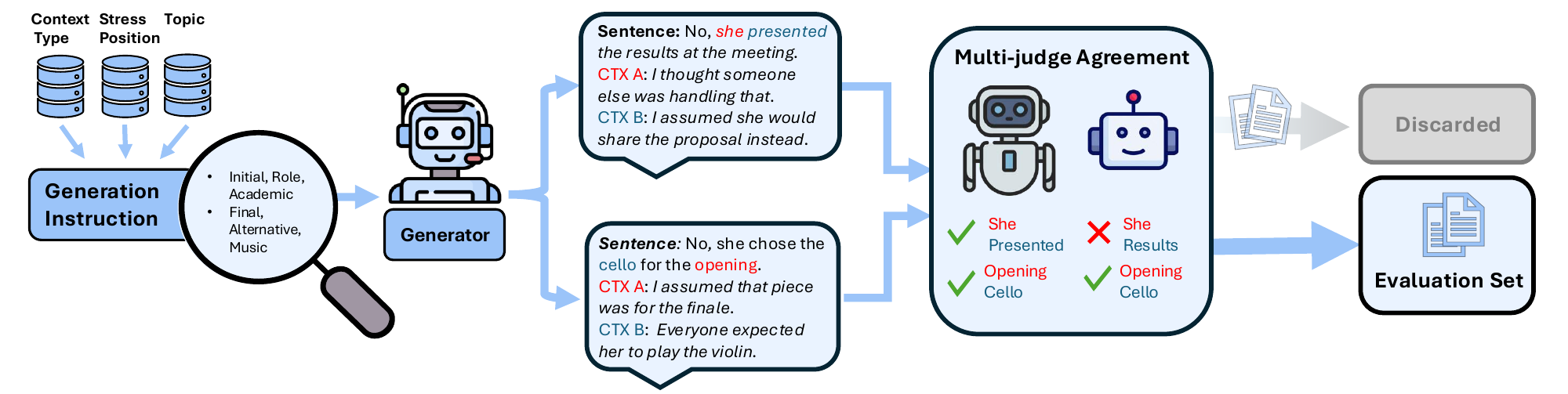}
   \caption{Overview of the \bench\ construction and validation pipeline. Contrastive context pairs are generated via structured prompting and filtered through multi-judge consistency checks.}
    \label{fig:fig2}
\end{figure*}
Several benchmarks evaluate stress or emphasis from different perspectives. Stress detection tasks derive prominence labels from speech corpora and frame the problem as predicting observed emphasis \cite{morrison2024crowdsourced}. The StressTest benchmark \cite{yosha2025stresstest} constructs contrastive examples to study stress understanding, but focuses on language understanding rather than speech generation. EmphAssess \cite{de2024emphassess} and Prosaudit \cite{de2023prosaudit} evaluate emphasis realization or transfer, measuring whether models preserve or reproduce explicitly marked stress. These evaluations assess stress detection or controlled realization, but do not test context-driven stress shifts in TTS generation.

Overall, existing work either models prosody given explicit control, detects stress from speech, or studies contrastive focus at the text level without speech synthesis. In contrast, \bench\ defines intended stress semantically and evaluates end-to-end TTS systems on their ability to infer and realize context-conditioned stress across controlled contrastive context pairs.

\section{\bench}
\label{sec:bench}

\subsection{Task}

We consider the task of context-conditioned stress generation in TTS. The input consists of a textual context $c$ and a target sentence $s = (w_1, \dots, w_n)$. The output is a synthesized speech signal $\hat{y}$ corresponding to $s$. No stress annotations or emphasis markers are provided at inference time.
Each benchmark item is defined by a tuple $(c, s, w^*)$, where $w^*$ denotes the intended stressed word in $s$, defined semantically based on the discourse context.
A successful system must infer the appropriate stress target from $\boldsymbol{c}$ and $\boldsymbol{s}$, and realize that emphasis acoustically in the generated speech. Table~\ref{tab:data_examples} illustrates contrastive context pairs in which the same sentence requires different stress targets depending on the preceding discourse. This contrastive design ensures that any observed stress difference can be attributed to context rather than sentence-internal biases, as the lexical content is held constant across conditions.

\begin{table}[t]
\centering
\caption{Examples from \bench. Each target sentence is paired with two contexts. The intended stressed word for Context A is shown in blue, and for Context B in red.}
\label{tab:data_examples}
\begin{tabular}{p{0.45\linewidth} p{0.45\linewidth}}
\toprule
\textbf{Context} & \textbf{Target Sentence} \\
\midrule

\stressA{A}: The editor wrote it ran in the spring. &
She \stressB{published} the essay in the \stressA{winter} issue. \\
\stressB{B}: I believed she sent a draft. & \\

\midrule

\stressA{A}: I thought Tom was the mentor here. & 
The incubator selected \stressA{Priya} as the startup \stressB{mentor}. \\
\stressB{B}: People assumed she would run the investment team. & \\

\bottomrule
\end{tabular}
\end{table}

\subsection{Benchmark Construction}
\label{sec:bench_const}
The \bench\ evaluation set is constructed entirely at the textual level. Contexts, sentences, and intended stress targets are generated jointly using structured prompting with \textit{gpt-5-mini}. Generating these elements together ensures semantic coherence: the context motivates a particular interpretation of the sentence, and the stressed word is selected as part of the same generative process rather than inferred post hoc.
Each sentence is constructed to contain exactly two plausible stress candidates. Generation is controlled via planning codes that specify the stress position and the discourse function of the first context variation (e.g., correction, role disambiguation, exclusivity). The second variation has more freedom, constrained only to stress the other candidate word. Full generation and sampling details are released alongside the pipeline.

Generated items are filtered through a multi-judge consistency check using two independent models: \textit{gpt-5-nano} and \textit{gemini-2.5-flash}. We discard items where the two judges disagree on the intended stressed word given the context. To mitigate potential circularity, \textit{gemini-2.5-flash} serves as an independent judge from a different model family than the generator. Human annotators confirmed label quality on a subset covering all stress positions and context types present in the full benchmark (Section~\ref{sec:human_val}). Items failing either check are discarded, yielding the final benchmark set.
We note that no speech synthesis or acoustic information is used at any stage of construction. During evaluation, stress correctness is assessed from speech produced by the evaluated system, not from reference audio. An overview of the construction pipeline is illustrated in Figure~\ref{fig:fig2}.

\begin{table}[t]
\centering
\caption{Diversity statistics of the \bench. The target stress position is balanced across the sentence, preventing models from exploiting position bias. Context types cover a wide range of pragmatic phenomena.}
\label{tab:dataset_stats}
{
\begin{tabular}{l c | l c}
\toprule
\textbf{Stress Position} & \textbf{\%} & \textbf{Context Type} & \textbf{\%} \\
\midrule
Initial     & 23.9 & Alternative  & 18.6 \\
Early       & 23.9 & Exclusivity  & 16.8 \\
Medial      & 23.0 & Role         & 15.9 \\
Final       & 29.2 & Correction   & 13.3 \\
            &      & Location     & 12.4 \\
            &      & Quantity     & 10.6 \\
            &      & Timing       & 7.1  \\
            &      & Modality     & 5.3  \\
\bottomrule
\end{tabular}
}
\end{table}

\subsection{Dataset Overview}
The \bench\ evaluation set contains 113 contrastive context pairs (226 items), validated through multi-judge agreement as described above.
Table~\ref{tab:dataset_stats} shows that the evaluation set is balanced across stress positions and pragmatic phenomena. Because stress targets are distributed across sentence positions and discourse types, models are unlikely to succeed through positional heuristics or shallow keyword matching.

\subsection{Training Resource}
In addition to the benchmark, we release an extended version of the construction pipeline that supports audio synthesis and automatic stress-based filtering. This enables scalable construction of context-sentence-audio triples for supervised training. We release an initial corpus of $\sim$10k samples generated by this pipeline to support future work on training context-aware TTS systems. This corpus is separate from the benchmark and is not used in evaluation.

\section{Experimental Setup}
\label{sec:setup}

\subsection{Evaluated Systems}
\label{sec:models}
We evaluate a diverse set of TTS systems, each under the conditioning modes it natively supports. In addition to baseline (no context), we consider three modes: (1) Concat: the context is prepended to the target sentence as a single input. The full audio is synthesized from the concatenated input, and the target sentence boundaries are identified using Whisper-based forced alignment. Evaluation metrics are computed only on the extracted segment, ensuring comparability across conditioning modes. (2) Instruct: the context is provided as a natural language instruction, e.g., \textit{"The following context determines which word should be emphasized in the sentence: [context]"}.
(3) Explicit: the target stressed word is specified using model-supported markup. \textbf{Kokoro} is a conventional non-LLM system, included as a reference for sentence-internal stress patterns. \textbf{Chatterbox}~\cite{chatterboxtts2025} is an LLM-based system without instruction support. \textbf{CosyVoice}, \textbf{HiggsAudio V2}, and \textbf{Qwen3-TTS}~\cite{CosyVoice,higgs,qwen3tts} are publicly available
instruction-capable LLM-based systems. Lastly, \textbf{GPT-4o-mini-tts} is a proprietary high-performance instruction-capable system, also evaluated under explicit-stress conditions. 

\begin{table}[t]
\centering
\caption{
Stress realization results (\%). Hit: target word stressed; Pair-Contrast: target present and alternative absent; Pair-Correct: both sides correct. Explicit stress is an oracle upper bound. $\pm=95\%$ bootstrap CI over 10K resamplings.
}
\label{tab:end_to_end}
\setlength{\tabcolsep}{4pt}
\renewcommand{\arraystretch}{0.97}
\begin{adjustbox}{width=\columnwidth}
\begin{tabular}{l l c c c}
\toprule
\textbf{Model} & \textbf{Stress}
& \textbf{Hit}
& \textbf{Contrast}
& \textbf{Correct} \\
\midrule
Kokoro
& -        & $38.1 \pm 6.4$ & $22.1 \pm 4.4$ & $0.0$ \\
\midrule
Chatterbox
& -        & $23.0 \pm 5.6$ & $16.8 \pm 4.4$ & $0.0$ \\
& Concat   & $22.6 \pm 5.8$ & $15.5 \pm 4.2$ & $0.0$ \\
\midrule
CosyVoice3
& -        & $32.3 \pm 5.8$ & $23.5 \pm 4.9$ & $0.9 \pm 1.4$ \\
& Concat   & $36.3 \pm 5.8$ & $23.9 \pm 4.7$ & $0.0$ \\
& Instruct & $32.3 \pm 6.0$ & $23.0 \pm 4.9$ & $0.9 \pm 1.4$ \\
& Explicit & $52.2 \pm 6.0$ & $40.3 \pm 5.8$ & $10.6 \pm 5.8$ \\
\midrule
GPT-4o-mini
& Instruct & $35.4 \pm 6.2$ & $26.1 \pm 5.3$ & $3.5 \pm 3.1$ \\
& Explicit & $51.8 \pm 6.2$ & $43.8 \pm 5.3$ & $11.5 \pm 5.8$ \\
\midrule
HiggsAudio V2
& Instruct & $30.1 \pm 6.0$ & $23.0 \pm 4.9$ & $2.7 \pm 3.1$ \\
\midrule
Qwen3TTS
& -        & $42.5 \pm 6.0$ & $26.1 \pm 5.1$ & $2.7 \pm 3.1$ \\
& Concat   & $35.0 \pm 6.0$ & $24.8 \pm 4.9$ & $2.7 \pm 3.1$ \\
& Instruct & $38.9 \pm 6.5$ & $22.6 \pm 5.3$ & $3.5 \pm 3.1$ \\
\bottomrule
\end{tabular}
\end{adjustbox}
\end{table}

\subsection{Evaluation Protocol}
\label{sec:evaluation}

All systems are evaluated exclusively based on their generated audio, without access 
to reference speech or oracle stress annotations at inference time. The objective is 
to determine whether synthesized speech (i) realizes the intended stressed word and 
(ii) adapts stress appropriately across contrasting contexts.

\noindent\textbf{Stress Detection.}
To enable scalable evaluation across systems and input conditions, we employ \textsc{WhiStress}~\cite{yosha2025whistress}, an automatic stress detection model built upon Whisper and fine-tuned to identify prominent stressed words in spoken utterances. Given synthesized audio $\hat{y}$ for a sentence 
$s$, the detector outputs a set of predicted stressed words $\hat{Y}$, applied uniformly across all evaluated systems. The evaluation protocol is detector-agnostic and can accommodate future stress detection models. We validate \textsc{WhiStress} as a reliable proxy for human stress perception in Section~\ref{sec:human_val}.

\noindent\textbf{Stress Metrics.}
Standard precision, recall, and F1 measure whether the intended word is stressed, but can miss context sensitivity. A system that consistently stresses the same word regardless of context could achieve high F1 while failing the core task. We therefore introduce contrastive metrics that explicitly measure adaptation across contexts. Let $A$ and $B$ denote the intended stressed words for a contrastive context 
pair (as defined in Section~\ref{sec:bench}), and let $\hat{Y}_A$ and $\hat{Y}_B$ 
denote the detected stressed words for each synthesized output. 

\textbf{Hit:} a per-sample metric indicating whether the intended stressed word appears among detected stressed words: \[\mathbf{1}[A \in \hat{Y_A}]\]and symmetrically for $(c_B, s,B)$. This measures basic stress expressiveness, regardless of additional stressed words.

\textbf{Pair-Contrast (Contrast):} a per-sample metric requiring the correct stressed word to be present \emph{and} the alternative context's stressed word to be absent:
\[
\mathbf{1}\big[ A \in \hat{Y}_A \ \wedge\ B \notin \hat{Y}_A \big]
\]
and symmetrically for $(c_B, s)$. This goes beyond Hit by penalizing over-expressive stress: the intended word must be stressed while the word that would be correct under the opposing context must not. A system that stresses too many words will fail here. However, a system with a sentence-internal bias toward one candidate may still satisfy Pair Contrast for one side of the pair by chance.

\textbf{Pair-Correct (Correct):} a pair-level metric requiring Pair Contrast to hold for \emph{both} sides of the contrastive context pair:
\[
\mathbf{1}\Big[
(A \in \hat{Y}_A \wedge B \notin \hat{Y}_A)
\ \wedge\
(B \in \hat{Y}_B \wedge A \notin \hat{Y}_B)
\Big].
\]
This is the strictest metric, requiring Pair Contrast to hold for both sides simultaneously. Since the two samples share the same sentence, any sentence-internal bias will produce the same stress pattern for both, making it impossible to satisfy Pair Correct without genuinely responding to the differing contexts.

\subsection{Human Validation}
\label{sec:human_val}

We conduct two human evaluation studies to validate (1)~the benchmark's 
ground-truth labels and (2)~the automatic stress detector used for evaluation.

\noindent\textbf{Benchmark Labels.}
To verify that the LLM-generated contrastive context stress labels reflect human judgment,
three annotators, all fluent English speakers, independently identified the contextually stressed word in 50~items 
(25~contrastive context pairs). Individual accuracy against benchmark labels ranged from 
$80\%$ to $94\%$, with average pairwise agreement of $79\%$. A majority consensus 
($\geq$2/3) was reached on $94\%$ of items, matching the ground truth in $93.6\%$ 
of those cases (Pair-Correct: $95.5\%$). The three items where the majority diverged 
from the benchmark involved sentences with multiple plausible focal words, confirming 
that the generated labels are well-calibrated with human intuition about contrastive context
stress.

\noindent\textbf{Stress Detector.}
To assess whether \textsc{WhiStress} serves as a reliable proxy for human stress 
perception, three listeners, all fluent English speakers, annotated 50~synthesized utterances 
(25~CosyVoice explicit, 25~gpt-4o-mini-tts context-instruct), marking all words they 
perceived as stressed. Inter-annotator agreement was Fleiss'~$\kappa=0.36$, 
consistent with prior work on prosodic prominence perception~\cite{BISHOP2020100977}. Human--\textsc{WhiStress} agreement (majority-vote vs.\ detector, $\kappa=0.32$, 
$F1=0.48$) falls within the range of pairwise inter-annotator agreement 
($\kappa=0.29$--$0.40$), indicating that detector disagreements with any individual 
listener are no larger than disagreements among listeners themselves.

\section{Results}
\label{sec:results}

\subsection{Context-Aware Stress Realization}
\label{sec:end_to_end_results}

Table~\ref{tab:end_to_end} summarizes performance across models and input modes. Overall, all evaluated systems exhibit limited reliability in context-dependent stress realization. While several models achieve moderate Hit scores, indicating that the target word is sometimes stressed, Pair-Contrast and Pair-Correct remain substantially lower across all systems. Pair-Correct scores are near zero for nearly all systems, suggesting that models frequently default to sentence-internal stress patterns rather than adapting to discourse context.

Providing context yields no clear improvement improvement across any system or conditioning mode. Confidence intervals (CIs) for Pair-Contrast and Pair-Correct overlap substantially across all context conditions, indicating that current TTS systems do not effectively utilize context regardless of how it is provided.

Explicit stress supervision provides a clearly higher upper bound, with 
non-overlapping CIs relative to all context conditions. CosyVoice with explicit stress reaches Pair-Contrast of $40.3 \pm 5.8$ and Pair Correct of $10.6 \pm 5.8$, suggesting that the capacity for stress realization exists but is not reliably activated by contextual input alone. Nevertheless, even explicit supervision falls short of reliable realization and remains a challenge itself.

\subsection{Text-Only Stress Inference}
\label{sec:text_only}

To isolate semantic stress inference from acoustic realization, we evaluate text-only language models on the same benchmark. Models receive conversational context and sentence text and directly predict the intended stressed word. All models are evaluated zero-shot, without in-context examples. For models that support a thinking mode, we use the standard non-thinking.
Table~\ref{tab:text_only_results_full} shows that text-only models substantially outperform all end-to-end TTS systems. Performance improves consistently with model scale, with Qwen3 models achieving Contrast up to 57.5\%. Claude Haiku achieves 88.1\% Contrast and 76.1\% Correct. These results confirm that the intended stress is largely recoverable from context at the text level, establishing a strong upper bound for speech generation. We note that while the evaluated text models may predict incorrectly, they consistently predict exactly one stressed word per item. As a result, Hit and Contrast are identical.

\begin{table}[t]
\centering
\caption{Text-only stress inference (\%). Models predict the stressed word directly from context and sentence.}
\label{tab:text_only_results_full}
\begin{tabular}{l c c c}
\toprule
\textbf{Model} & \textbf{Hit} $\uparrow$ & \textbf{Contrast} $\uparrow$ & \textbf{Correct} $\uparrow$ \\
\midrule
Qwen3-0.6B     & 25.2 & 25.2 & 5.3 \\
Qwen3-1.7B     & 50.0 & 50.0 & 24.8 \\
Qwen3-4B       & 57.5 & 57.5 & 37.2 \\
Claude-haiku-4.5 & 88.1 & 88.1 & 76.1 \\
\bottomrule
\end{tabular}
\end{table}

\section{Discussion}
\label{sec:discussion}

Our results reveal a consistent gap between text-level stress inference and speech-level stress realization in current TTS systems. Text-only models demonstrate that the intended stressed word is largely recoverable from discourse context, yet TTS systems fail to reliably produce context-appropriate stress in speech. This gap persists across all evaluated systems and conditioning modes. The explicit stress oracle, where the target word is directly specified, shows better results, though it remains far from perfect, suggesting room for improvement in both stress realization and contextual understanding.

Our evaluation relies on \textsc{WhiStress} as the stress detector, which may favor certain acoustic cues such as loudness over subtler pitch-based emphasis. We validate its reliability against human perception in Section~\ref{sec:human_val}, but acknowledge that automatic detection remains an imperfect proxy.

Additionally, we focus on lexical stress as a controlled and measurable starting point. Broader prosodic dimensions such as intonation contour and phrasing may also carry contextual intent and represent a natural direction for future work.

\section{Conclusion}
\label{sec:conclusion}

We introduced \bench, a benchmark for evaluating context-conditioned word-level stress in TTS. By defining intended stress semantically through contrastive context pairs and evaluating realization directly from synthesized speech, the benchmark isolates a core open challenge in expressive TTS.

Our results reveal a consistent gap: while the intended stressed word is largely recoverable from text, current TTS systems fail to realize this in speech across all evaluated architectures and conditioning modes. We hope future work will advance not only explicit stress realization but also the ability of TTS systems to infer appropriate stress from discourse context.

We release the benchmark, evaluation framework, construction pipeline, and a synthetic training corpus to support research on context-aware stress in speech synthesis.

\section{Generative AI Use Disclosure}
Generative AI tools were used as part of the research methodology, as described in Section~\ref{sec:bench_const}, and for editing and polishing the manuscript text.

\bibliographystyle{IEEEtran}
\bibliography{mybib}

\end{document}